\documentclass[10pt,twocolumn,letterpaper]{article}

\usepackage{cvpr}
\usepackage{times}
\usepackage{epsfig}
\usepackage{graphicx}
\usepackage{amsmath}
\usepackage{amssymb}
\usepackage{subcaption}

\usepackage[breaklinks=true,bookmarks=false]{hyperref}

\cvprfinalcopy % *** Uncomment this line for the final submission

\def\cvprPaperID{****} % *** Enter the CVPR Paper ID here

\begin{document}

%%%%%%%%% TITLE
\title{Understanding the visual speech signal}

\author{Helen L. Bear\\
University of East London, 4-6 University Way, London E16 2RD\\
{\tt\small helen@uel.ac.uk}
}

\maketitle
%\thispagestyle{empty}

%%%%%%%%% ABSTRACT
\begin{abstract}
For machines to lipread, or understand speech from lip movement, they decode lip-motions (known as visemes) into the spoken sounds. 
We investigate the visual speech channel to further our understanding of visemes. This has applications beyond machine lipreading; speech therapists, animators, and psychologists can benefit from this work. We explain the influence of speaker individuality, and demonstrate how one can use visemes to boost lipreading.   	
\end{abstract}

%%%%%%%%% BODY TEXT
\section{Introduction}
Machine lipreading (MLR) is speech recognition without audio input \eg from a silent video. MLR research is of interest to computer vision engineers and speech researchers. Two current complimentary challenges in MLR are; to develop an end-to end system or, to understand the visual speech signal to apply the knowledge to new domains such as speech therapy and animation. Our work addresses the latter challenge. 
Phonemes are the smallest sounds one can make \cite{international1999handbook}, and a viseme is the visual equivalent \cite{7074217}. Current knowledge of visemes is limited, there is no proven function, (often presented as a map) between visemes and phonemes. Our work here focuses on understanding visemes, in order to recognise the right phoneme. 

\subsection{Conventional lipreading machines}
The conventional lipreading process has, at a high level, been adopted from audio recognition systems. This is: 1) track faces and extract features, 2) train a model and classify 3) filter output through a language network. Debates over the optimal tracking methods, features \cite{cappelletta2012phoneme}, and classifier method \cite{shin2011real} remain but, pre-deep learning, the classic choices with accurate results were Active Appearance Model features \cite{Matthews_Baker_2004} and Hidden Markov Model classifiers \cite{rabiner1986introduction} (often built with the HTK toolkit \cite{young2006htk} \eg \cite{lee2002audio,puviarasan2011lip,saitoh2008analysis}). 

\subsection{Data} 
Available lipreading datasets are reviewed in \cite{bear_harvey2} but the most accurate lipreading data to date are; BBC \cite{chungaccv}, TCD-TIMIT \cite{7050271}, Oulu \cite{7163155}, and RMAV \cite{improveVis}. We use RMAV.

\section{The phoneme-to-viseme map play-off}
\begin{figure}[h]
\centering
   \includegraphics[width=0.9\linewidth]{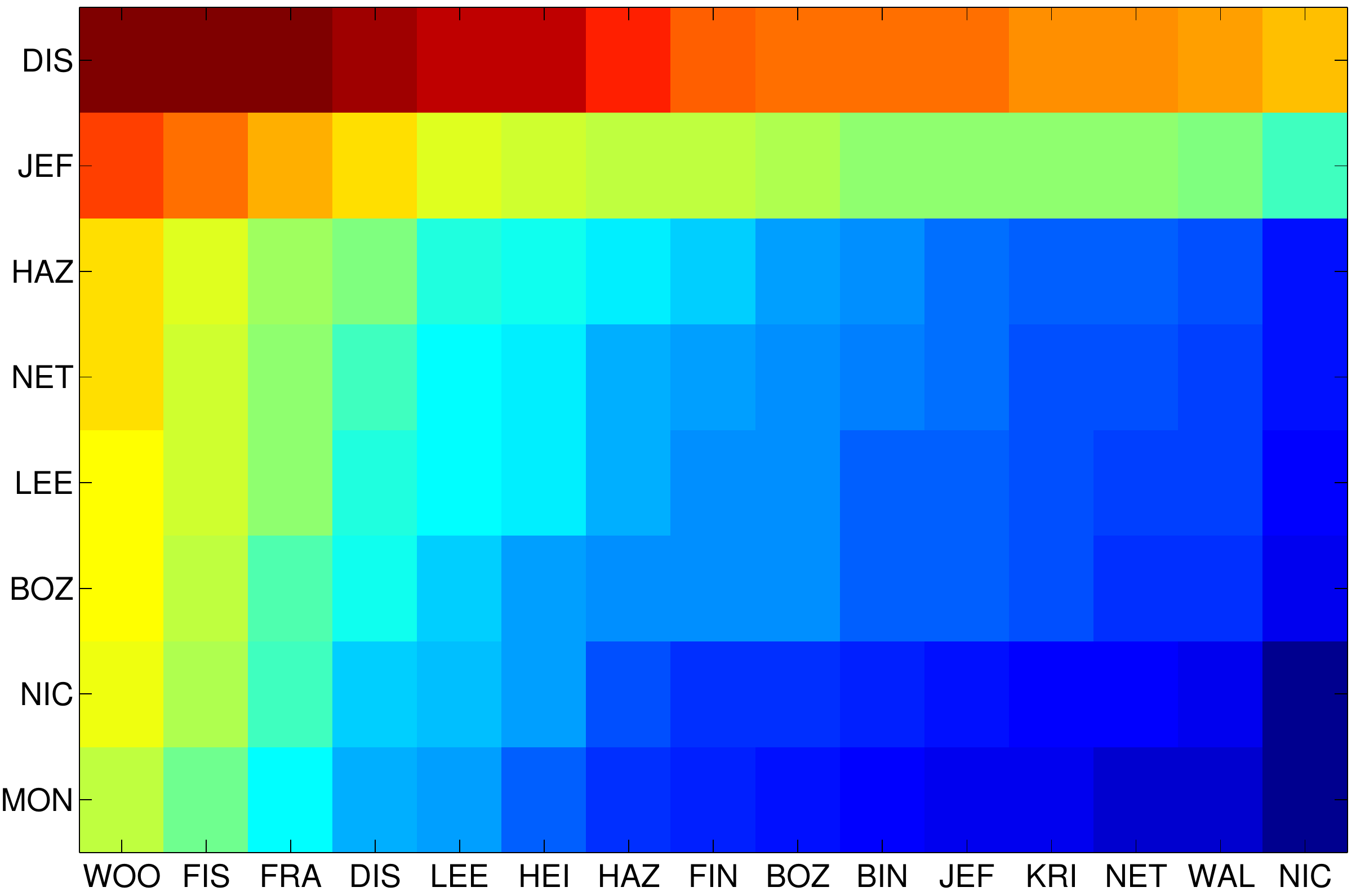}
   \caption{Heatmap of lipreading P2V maps.\cite{bear_harvey1}}
   \label{fig:heatmapISVC}
\end{figure}

We begin with a play-off to measure the effect of using different phoneme-to-viseme (P2V) maps from prior work. $120$ P2Vs are tested with the conventional system on $12$ talkers. The results are displayed in Figure~\ref{fig:heatmapISVC} as a heatmap \cite{bear_harvey1}. Consonant P2Vs are on the $x$-axis and vowel P2Vs on the $y$-axis. We see that a combination of Disney vowels \cite{disney} and Woodward consonants \cite{woodward1960phoneme} perform best. This contrasts with \cite{bear2014phoneme} which concluded Lee's visemes \cite{lee2002audio} achieved most accurate lipreading with isolated words which suggests that utterance duration affects visemes.

\begin{figure}[h]
\centering
\begin{subfigure}[b]{0.25\textwidth}
   \includegraphics[width=0.95\linewidth]{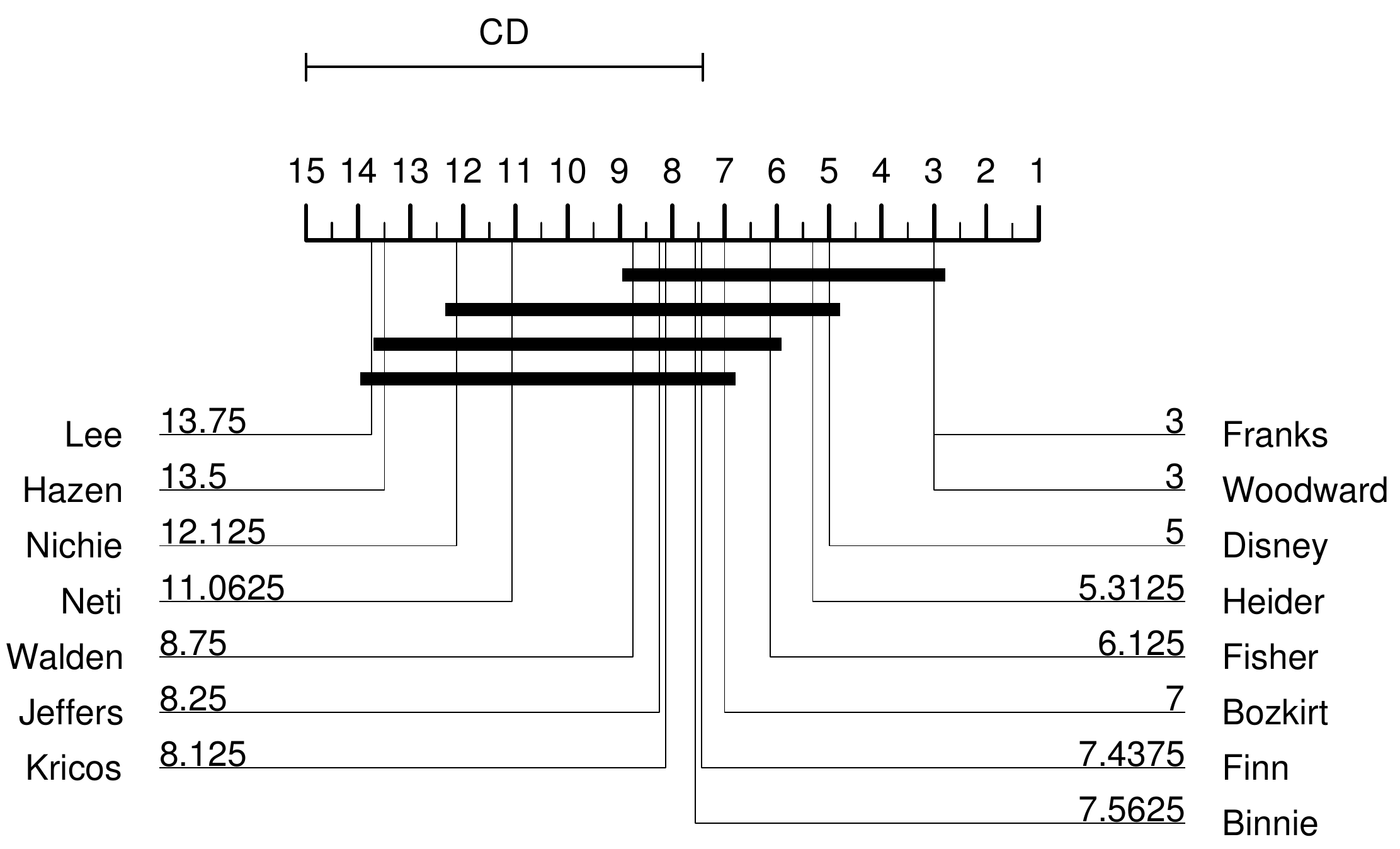}
   \caption{Consonant P2V maps}
    \label{fig:critDiffC}
\end{subfigure}%
\begin{subfigure}[b]{0.25\textwidth}
   \includegraphics[width=0.95\linewidth]{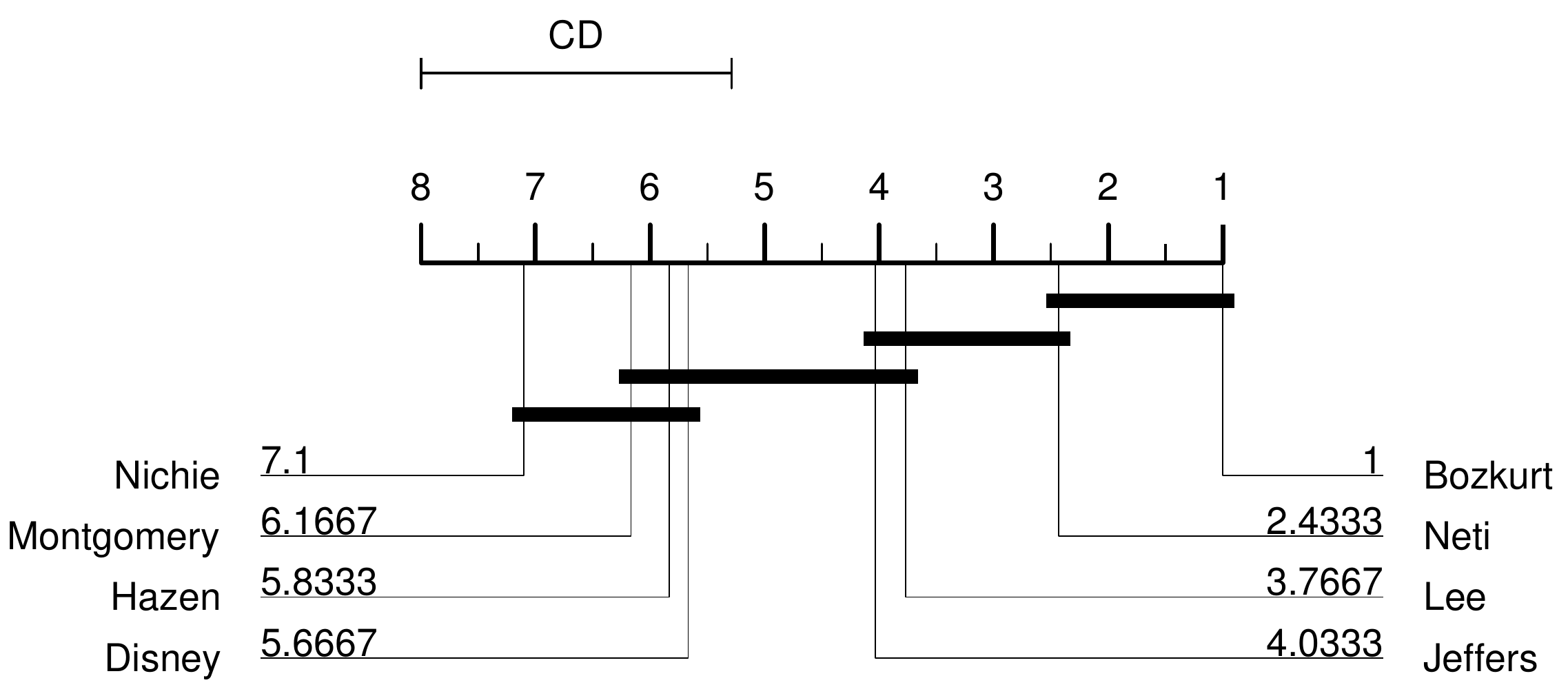}
   \caption{Vowel P2V maps.}
    \label{fig:critDiffV}
\end{subfigure}%
\caption{Critical difference between P2V maps.}
\label{fig:crit_diff}
\end{figure}

Figure~\ref{fig:crit_diff} is critical difference plots for the P2V maps. Critical difference is a measure of confidence intervals between different algorithms \cite{criticaldiff}. Overlapping bars join P2V maps which are not critically different. By comparing Figures~\ref{fig:critDiffC} and~\ref{fig:critDiffV} we see that consonant visemes vary less than the vowel sets. This observation is supported by lipreading practitioners (\eg Nichie \cite{nrrchm1912lipreading}), who advocate there are key shapes for articulator sounds (vowels) and gestures are formed by motion between the shapes, the motions are determined by consonants. 

All P2V maps are fully tabulated in \cite{yogiPhd,bear2014phoneme}. 
\section{Speaker independence}
\begin{figure}[h]
\centering
   \includegraphics[width=0.9\linewidth]{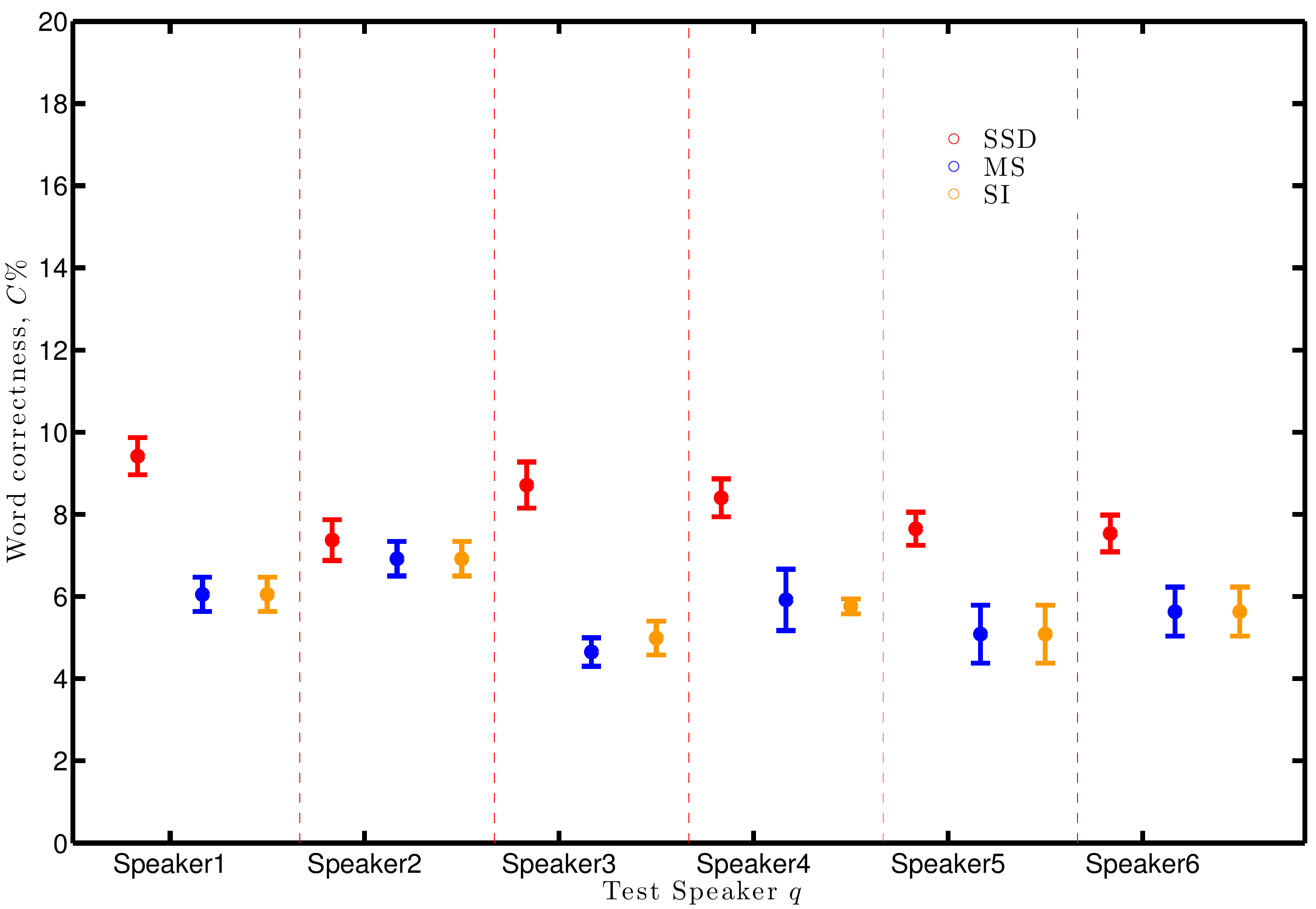}
   \caption{Comparing multi-speaker, speaker-dependent P2V functions on six RMAV speakers.}
\label{fig:heatmapSpeakIndep}
\end{figure}

In \cite{bear2015speakerindep} results show Speaker-Dependent (SD) visemes can improve lipreading accuracy. In Figure~\ref{fig:heatmapSpeakIndep} this conclusion is reinforced with equivalent experiments on continuous speech talkers. Red plots show SD visemes, blue plots are Multi-speaker (MS) visemes, and orange are Speaker-Independent (SI) visemes. Speaker independence is the ability to lipread previously unseen talkers and is an obstacle for lipreading machines. 

In \cite{bear2014some} we learn there is a limitation on how useful all SI visemes within a set are towards recognition accuracy. A badly trained viseme is worse than no viseme. However with our SD visemes, (red plots in Figure~\ref{fig:heatmapSpeakIndep}) all visemes increase accuracy. So, whilst bad training data is more detrimental to classification than having less, with the right knowledge of visual gestures, our need for big data is reduced for accurate lipreading. 

\section{Boosting phonemes with visemes}
\begin{figure}[h]
\centering
   \includegraphics[width=0.9\linewidth]{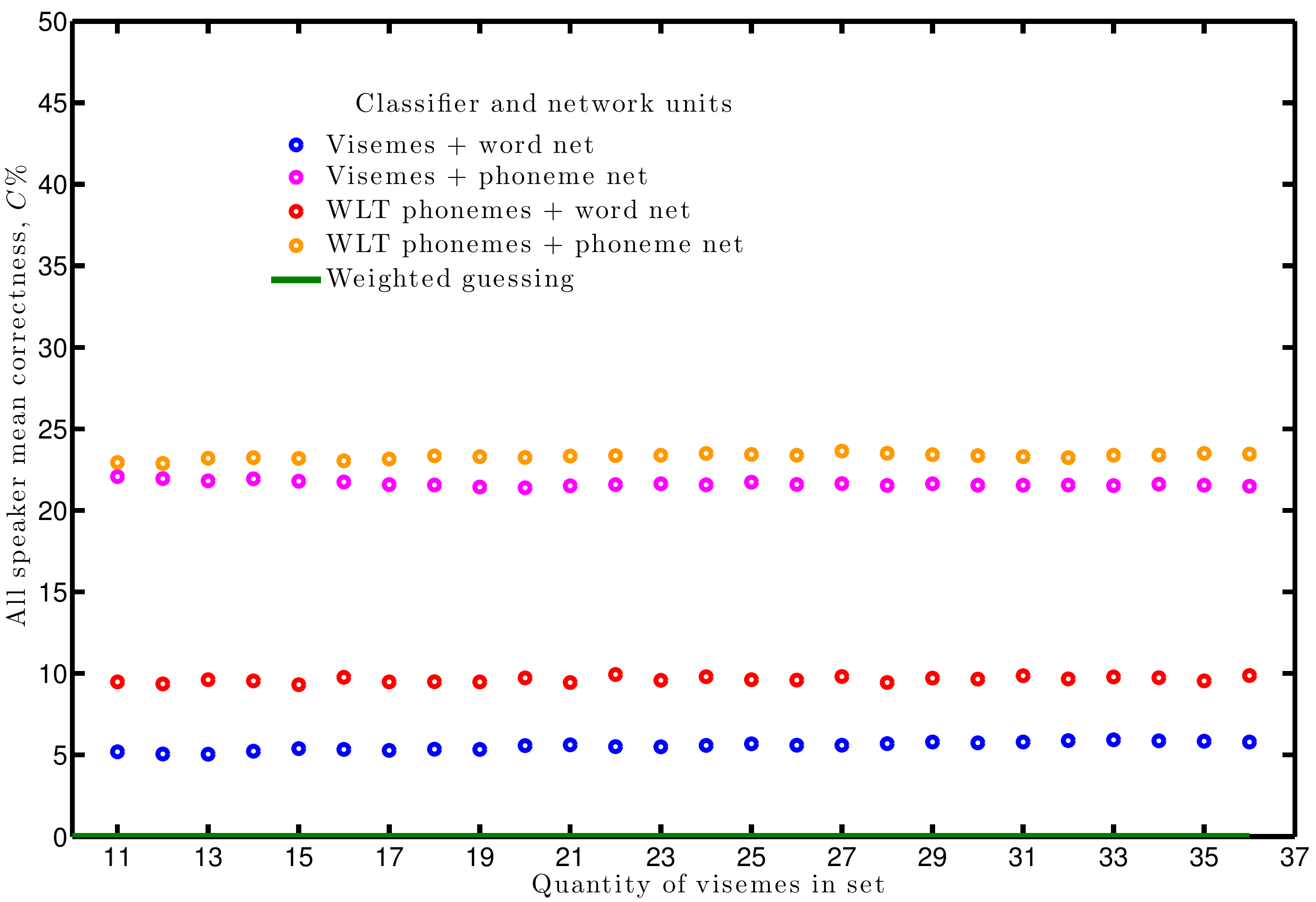}
   \caption{Boosting with network decoding and classifier units.}
\label{fig:heatmapICASSP}
\end{figure}

We present an experiment in \cite{bear2015findingphonemes} which showed viseme sets with $<11$ visemes are negatively affected by homophone confusions. The sets which are too large ($>35$) do not differentiate sufficiently to for accurate lipreading. This means the range of optimum sizes is from $11$ to $35$ and varies by talker. Further to this, in \cite{bear2016decoding} we designed a hierarchical training method which used viseme classifiers as initialisation models of phoneme classifiers, for all viseme set sizes. All talker mean results are in Figure~\ref{fig:heatmapICASSP}. Phoneme HMMs initialised with visemes achieve higher accuracy. 

We also tested the of the language network unit. In Figure~\ref{fig:heatmapICASSP} we show that a phoneme network is better than a word network. However, using a phoneme network means the final output is a phoneme string which requires further processing to understand but in \cite{bear2015findingphonemes} this effect is not significant.

\section{Conclusions and the future}
In our comparison of previous P2V mappings there is little difference between them but Disney's outperforms others on continuous speech and Lee's marginally outperforms others \cite{bear2014phoneme} on isolated words. This means that visemes vary, by speaker and, by utterance. We suggest that speaker individuality in visual speech is due to the variability with which different people use visual gestures whilst talking. 

For speaker-dependent recognition there are choices when selecting a set of visemes containing fewer classes than the phoneme set, yet these sets outperform phoneme labelled classifiers. But phoneme classifiers are desirable as these are cross-speaker consistent so we ask is there a way of mapping similarities between SD visemes \cite{bear2017function}? For not only can the right SD visemes out-perform phoneme classifiers, but when used to help train phoneme classifiers, they lipread significantly better \cite{bear2016decoding} also. 

Best results are achieved when the units match between classifiers and the language network, but not significantly so. So, for the purposes of decoding phonemes to the words spoken, the preferred network unit is words \cite{bear2016decoding}.

End-to-end systems perform well with big data and deep learning \cite{chungaccv} but we are still to fully understand the visual speech signal. Understanding visual speech will mean we can improve adaptation between talkers in the future.

{\small
\bibliographystyle{ieee}
\bibliography{egbib}

\begin{thebibliography}{10}\itemsep=-1pt

\bibitem{7163155}
I.~Anina, Z.~Zhou, G.~Zhao, and M.~PietikŠinen.
\newblock Ouluvs2: A multi-view audiovisual database for non-rigid mouth motion
  analysis.
\newblock In {\em 2015 11th IEEE International Conference and Workshops on
  Automatic Face and Gesture Recognition (FG)}, volume~1, pages 1--5, May 2015.

\bibitem{international1999handbook}
I.~P. Association.
\newblock {\em Handbook of the International Phonetic Association: A guide to
  the use of the International Phonetic Alphabet}.
\newblock Cambridge University Press, 1999.

\bibitem{bear_harvey2}
Authors.
\newblock When will machine lipreading come of age? have we reached a
  singularity?
\newblock {\em Signal Processing Magazine - under review}, 2017.

\bibitem{yogiPhd}
H.~L. Bear.
\newblock Decoding visemes: improving machine lipreading, 2016.

\bibitem{bear2017function}
H.~L. Bear.
\newblock Visual gesture variability between talkers in continuous visual
  speech.
\newblock In {\em British Machine Vision Conference (BMVC) Deep learning for
  machine lip reading workshop}. British Machine Vision Association (BMVA),
  2017.

\bibitem{bear2015speakerindep}
H.~L. Bear, S.~J. Cox, and R.~Harvey.
\newblock Speaker independent machine lip reading with speaker dependent viseme
  classifiers.
\newblock In {\em 1st Joint International Conference on Facial Analysis,
  Animation and Audio-Visual Speech Processing (FAAVSP)}, pages 115--120. ISCA,
  2015.

\bibitem{bear2016decoding}
H.~L. Bear and R.~Harvey.
\newblock Decoding visemes: improving machine lip-reading.
\newblock In {\em 41st International Conference Acoustics, Speech and Signal
  Processing (ICASSP)}, 2016.

\bibitem{bear_harvey1}
H.~L. Bear and R.~Harvey.
\newblock Phoneme-to-viseme mappings: the good, the bad, and the ugly.
\newblock {\em Speech Communication: special issue on auditory-visual
  expressive speech}, 2017.

\bibitem{bear2015findingphonemes}
H.~L. Bear, R.~Harvey, B.-J. Theobald, and Y.~Lan.
\newblock Finding phonemes: improving machine lip-reading.
\newblock In {\em 1st Joint International Conference on Facial Analysis,
  Animation and Audio-Visual Speech Processing (FAAVSP)}, pages 190--195. ISCA,
  2015.

\bibitem{bear2014phoneme}
H.~L. Bear, R.~W. Harvey, B.-J. Theobald, and Y.~Lan.
\newblock Which phoneme-to-viseme maps best improve visual-only computer
  lip-reading?
\newblock In {\em Advances in Visual Computing}, pages 230--239. Springer,
  2014.

\bibitem{bear2014some}
H.~L. Bear, G.~Owen, R.~Harvey, and B.-J. Theobald.
\newblock Some observations on computer lip-reading: moving from the dream to
  the reality.
\newblock In {\em SPIE Security+ Defence}, pages 92530G--92530G. International
  Society for Optics and Photonics, 2014.

\bibitem{7074217}
L.~Cappelletta and N.~Harte.
\newblock Viseme definitions comparison for visual-only speech recognition.
\newblock In {\em Signal Processing Conference, 2011 19th European}, pages
  2109--2113, Aug 2011.

\bibitem{cappelletta2012phoneme}
L.~Cappelletta and N.~Harte.
\newblock Phoneme-to-viseme mapping for visual speech recognition.
\newblock In {\em International Conference on Pattern Recognition Applications
  and Methods (ICPRAM)}, pages 322--329, 2012.

\bibitem{chungaccv}
J.~S. Chung and A.~Zisserman.
\newblock Lip reading in the wild.
\newblock In {\em Asian Conference on Computer Vision}, 2016.

\bibitem{criticaldiff}
J.~Demar.
\newblock Statistical comparisons of classifiers over multiple datasets.
\newblock {\em Journal of Machine Learning Research}, 7:1--30, 2006.

\bibitem{7050271}
N.~Harte and E.~Gillen.
\newblock Tcd-timit: An audio-visual corpus of continuous speech.
\newblock {\em IEEE Transactions on Multimedia}, 17(5):603--615, May 2015.

\bibitem{improveVis}
Y.~Lan, B.-J. Theobald, R.~Harvey, E.-J. Ong, and R.~Bowden.
\newblock Improving visual features for lip-reading.
\newblock {\em Proceedings of the International Conference on Audio-Visual
  Speech Processing (AVSP)}, 7(3):42--48, 2010.

\bibitem{disney}
J.~Lander.
\newblock Read my lips: Facial animation techniques.
\newblock
  \url{http://www.gamasutra.com/view/feature/131587/read_my_lips_facial_animation_.php}.
\newblock Accessed: 2014-01-28.

\bibitem{lee2002audio}
S.~Lee and D.~Yook.
\newblock Audio-to-visual conversion using hidden markov models.
\newblock In {\em Proceedings of Pacific Rim International Conference on
  Artificial Intelligence (PRICAI)}, pages 563--570. Springer, 2002.

\bibitem{Matthews_Baker_2004}
I.~Matthews and S.~Baker.
\newblock Active appearance models revisited.
\newblock {\em International Journal of Computer Vision}, 60(2):135--164, 2004.

\bibitem{nrrchm1912lipreading}
E.~Nichie.
\newblock Lipreading principles and practice, 1912.

\bibitem{puviarasan2011lip}
N.~Puviarasan and S.~Palanivel.
\newblock Lip reading of hearing impaired persons using hmm.
\newblock {\em Expert Systems with Applications}, 38(4):4477--4481, 2011.

\bibitem{rabiner1986introduction}
L.~Rabiner and B.~Juang.
\newblock An introduction to hidden markov models.
\newblock {\em ieee assp magazine}, 3(1):4--16, 1986.

\bibitem{saitoh2008analysis}
T.~Saitoh, K.~Morishita, and R.~Konishi.
\newblock Analysis of efficient lip reading method for various languages.
\newblock In {\em Pattern Recognition, 2008. ICPR 2008. 19th International
  Conference on}, pages 1--4. IEEE, 2008.

\bibitem{shin2011real}
J.~Shin, J.~Lee, and D.~Kim.
\newblock Real-time lip reading system for isolated korean word recognition.
\newblock {\em Pattern Recognition}, 44(3):559--571, 2011.

\bibitem{woodward1960phoneme}
M.~F. Woodward and C.~G. Barber.
\newblock Phoneme perception in lipreading.
\newblock {\em Journal of Speech, Language and Hearing Research}, 3(3):212,
  1960.

\bibitem{young2006htk}
S.~J. Young, G.~Evermann, M.~Gales, D.~Kershaw, G.~Moore, J.~Odell, D.~Ollason,
  D.~Povey, V.~Valtchev, and P.~Woodland.
\newblock The {HTK} book version 3.4, 2006.

\end{thebibliography}
}

\end{document}